\documentclass{article} 
\usepackage{iclr2026_conference,times}

\usepackage{hyperref}
\usepackage{url}

\usepackage{amsmath,amssymb,amsfonts}
\usepackage{graphicx}
\usepackage{booktabs}
\usepackage{multirow}
\usepackage{enumitem}
\usepackage{xcolor}
\usepackage{caption}
\usepackage{subcaption}

\usepackage{algorithm}
\usepackage{algpseudocode}

\newcommand{\res}{r}

\newcommand{\icset}{\mathcal{I}}
\newcommand{\bcset}{\mathcal{B}}
\newcommand{\candset}{\mathcal{C}}
\newcommand{\selset}{\mathcal{S}}

\newcommand{\maybeincludegraphics}[2][]{%
  \IfFileExists{#2}{\includegraphics[#1]{#2}}{\fbox{\parbox[c][1.2in][c]{0.95\linewidth}{\centering Missing file: \texttt{#2}}}}%
}

\title{Diversity-Aware Adaptive Collocation for Physics-Informed Neural Networks\\
via Sparse QUBO Optimization and Hybrid Coresets}

\author{%
Hadi Salloum \\
Research Center for Artificial Intelligence\\ Innopolis University, Innopolis, Russia \\
Moscow Center for Advanced Studies\\ 20, Kulakova Str., Moscow,
Russia\\
\texttt{h.salloum@innopolis.ru}
\And
Maximilian Mifsud Bonici\\
Laboratory of Quantum Computing\\ Innopolis University, Innopolis, Russia
Q-Deep \\
Kazan, Russia\\
\texttt{m.mifsudbonici@qdeep.net}
\And
Sinan Ibrahim \\
Center for Engineering Systems and Sciences \\ Moscow, Russia\\
Laboratory of Quantum Computing\\ Innopolis University, Innopolis, Russia\\
\texttt{s.ibrahim@rcdei.com}
\And
Pavel Osinenko \\
 Central University \\ and Center for Engineering Systems and Sciences\\ Moscow, Russia \\
 \texttt{p.osinenko@gmail.com}
\And
Alexei Kornaev \\
The Center for Top-Level Educational Programs in AI\\ Innopolis University, Innopolis, Russia \\
Research Center for Artificial Intelligence \\ National Medical Research Center of Oncology, Moscow, Russia 
\\ \texttt{a.kornaev@innopolis.ru}
}

\iclrfinalcopy 
\begin{document}

\maketitle

\begin{abstract}
Physics-Informed Neural Networks (PINNs) enforce governing equations by penalizing PDE residuals at interior collocation points, but standard collocation strategies-uniform sampling and residual-based adaptive refinement-can oversample smooth regions, produce highly correlated point sets, and incur unnecessary training cost. We reinterpret collocation selection as a coreset construction problem: from a large candidate pool, select a fixed-size subset that is simultaneously informative (high expected impact on reducing PDE error) and diverse (low redundancy under a space--time similarity notion). We formulate this as a QUBO/BQM objective with linear terms encoding residual-based importance and quadratic terms discouraging redundant selections. To avoid the scalability issues of dense k-hot QUBOs, we propose a sparse graph-based BQM built on a kNN similarity graph and an efficient repair procedure that enforces an exact collocation budget. We further introduce hybrid coverage anchors to guarantee global PDE enforcement. We evaluate on the 1D time-dependent viscous Burgers' equation with shock formation and report both accuracy and end-to-end time-to-accuracy, including a timing breakdown of selection overhead. Results demonstrate that sparse and hybrid formulations reduce selection overhead relative to dense QUBOs while matching or improving accuracy at fixed collocation budgets.
\end{abstract}

\section{Introduction}
Physics-Informed Neural Networks (PINNs) embed differential equation constraints into training objectives, offering a flexible framework for solving forward and inverse problems governed by partial differential equations (PDEs) \citep{raissi2019pinn, karniadakis2021piml}. Despite strong promise, practical PINN training often hinges on collocation strategy: which interior points should be used to evaluate and penalize the PDE residual \citep{wang2021gradient, wang2022ntk}. Uniform sampling is simple but frequently inefficient, especially in problems with localized structures (e.g., shocks, boundary layers, sharp fronts), where informative regions occupy a small fraction of the domain. Residual-based adaptive refinement \citep{lu2021deepxde, wu2023comprehensive} improves focus but can still generate highly correlated point sets and can neglect global coverage, potentially harming stability and generalization.

This work studies \emph{combinatorial, diversity-aware} collocation selection. Instead of only repeatedly adding points at high-residual locations, we explicitly seek a compact subset of interior collocation points that maximizes informational value while minimizing redundancy. The core idea parallels coreset construction in machine learning \citep{feldman2020coresets, mirzasoleiman2020coresets}: select fewer training constraints without sacrificing solution quality.

\paragraph{Contributions.}
We present a QUBO/BQM-based framework for collocation selection and several practical variants designed for real PINN workloads:
\begin{itemize}[leftmargin=1.1em, itemsep=0.15em]
\item \textbf{QUBO formulation for collocation coresets:} linear terms encode point importance (e.g., PDE residual magnitude) while quadratic terms penalize redundant pairs via similarity weights.
\item \textbf{Sparse BQM + repair:} we avoid dense cardinality penalties (which produce all-to-all couplers) by solving a sparse BQM on a kNN similarity graph and applying an efficient repair/refinement step to enforce an exact collocation budget.
\item \textbf{Hybrid anchors for global coverage:} we reserve a fraction of collocation points as stratified ``coverage anchors'' and use QUBO to select the remainder, preventing overconcentration on localized regions.

\end{itemize}

\section{Related Work}
\paragraph{Physics-Informed Neural Networks.}
PINNs were popularized as a unified approach for forward and inverse problems by Raissi \emph{et al.}~\citep{raissi2019pinn}; see \citet{karniadakis2021piml} for a comprehensive review. Subsequent work revealed critical training pathologies: \citet{wang2021gradient} identified gradient flow imbalances across loss terms, \citet{wang2022ntk} analyzed convergence failures through the neural tangent kernel, and \citet{krishnapriyan2021failure} characterized failure modes as PDE stiffness increases. Remedies include self-adaptive loss weighting \citep{mcclenny2023selfadaptive}, causal training schedules \citep{wang2024causal}, and Fourier feature embeddings to address spectral bias \citep{wang2021fourier}.

\paragraph{Adaptive collocation and residual refinement.}
Residual-based adaptive refinement (RAR) was introduced in DeepXDE \citep{lu2021deepxde} and has since been extended in several directions. \citet{wu2023comprehensive} provided a systematic comparison of non-adaptive and residual-based adaptive sampling strategies across thousands of configurations, establishing RAR-D and RAD as strong baselines. Importance sampling approaches reweight or resample collocation points proportional to the PDE residual \citep{nabian2021importance}, while deep generative models can learn residual-guided distributions \citep{tang2023daspinn}. Failure-informed sampling \citep{gao2023failure} casts adaptive collocation as a reliability problem, and the R3 method \citep{daw2023r3} addresses propagation failures via evolutionary point retention. Most recently, PINNACLE \citep{lau2024pinnacle} jointly optimizes all training point types using NTK-based criteria with greedy selection. However, purely residual-driven methods may still produce redundant point clusters and may under-enforce the PDE globally.

\paragraph{Coresets, diversity, and subset selection.}
Diversity-promoting selection appears across ML as facility location, k-center, determinantal point processes (DPPs), and submodular maximization. Coresets---compact weighted subsets that approximate full-data objectives---have been applied to efficient neural network training via submodular facility location \citep{mirzasoleiman2020coresets} and gradient matching \citep{killamsetty2021gradmatch}. The k-center formulation for active learning \citep{sener2018coreset} and submodular data selection \citep{wei2015submodularity} provide the geometric diversity perspective that motivates our approach. DPPs offer a principled repulsive sampling framework \citep{kulesza2012dpp}, with fixed-size variants \citep{kulesza2011kdpp} and applications to mini-batch diversification \citep{zhang2017dpp}. The classical greedy approximation guarantee for monotone submodular maximization \citep{nemhauser1978submodular} underpins many of these methods. Our approach focuses on QUBO/BQM formulations that naturally support both classical heuristics (simulated annealing) and quantum-ready solvers.

\paragraph{Conceptual comparison to adaptive and diversity-aware selection.}
Many PINN adaptive collocation methods are \emph{sequential} and \emph{importance-driven}: they repeatedly add or resample interior points where the current PDE residual is large, which improves focus but can lead to clustered, highly correlated constraints \cite{lu2021deepxde,wu2023comprehensive,nabian2021importance,gao2023failure,tang2023daspinn,daw2023r3}.
In contrast, our method treats interior collocation as a \emph{fixed-budget subset selection} problem: from a candidate pool we select exactly $K$ points by minimizing a binary quadratic objective that rewards residual-based utility while penalizing space--time redundancy via pairwise similarity on a sparse $k$NN graph, followed by an \emph{exact-$K$ repair} step enforcing the cardinality constraint.
Conceptually, this aligns with diversity-aware subset selection (coresets, submodular selection, and $k$-DPPs) \cite{sener2018coreset,wei2015submodularity,nemhauser1978submodular,krause2008sensor,kulesza2012dpp,kulesza2011kdpp}, but realizes the trade-off through an explicit QUBO/BQM energy minimized with annealing-style heuristics \cite{kochenberger2014qubo,kirkpatrick1983sa}.

\paragraph{QUBO optimization.}
QUBO is a standard form for combinatorial optimization \citep{kochenberger2014qubo} and is compatible with quantum annealing \citep{kadowaki1998quantum, johnson2011quantum} and classical approximate solvers such as simulated annealing \citep{kirkpatrick1983sa}. Practical QUBO modeling techniques are surveyed in \citet{glover2022qubo}, and Ising formulations for NP problems are catalogued by \citet{lucas2014ising}. We use D-Wave Ocean's classical simulated annealing sampler (\texttt{neal}) as a reproducible solver backend.

\section{Problem Setup}
\subsection{Governing Equation: Viscous Burgers' Equation}
We consider the 1D viscous Burgers' equation
\begin{equation}
u_t + u\,u_x = \nu\,u_{xx}, \quad x\in[-1,1], \quad t\in[0,T],
\label{eq:burgers}
\end{equation}
with standard initial and boundary conditions used in PINN benchmarks \citep{raissi2019pinn, basdevant1986burgers}. This problem is a canonical test case because shocks and steep gradients create localized regions of difficulty, making uniform collocation inefficient. The Cole--Hopf transformation \citep{cole1951burgers, hopf1950burgers} provides exact analytical solutions for verification.

\subsection{PINN Training Objective}
Let $u_\theta(x,t)$ be a neural network approximation. Define the PDE residual
\begin{equation}
\res(x,t) \;=\; \partial_t u_\theta(x,t) \;+\; u_\theta(x,t)\,\partial_x u_\theta(x,t) \;-\; \nu\,\partial_{xx}u_\theta(x,t).
\end{equation}
PINNs typically minimize a composite loss
\begin{equation}
\mathcal{J}(\theta)=
\lambda_{\mathrm{IC}}\;\mathbb{E}_{(x,t)\in\icset}\!\left[\bigl(u_\theta-u_{\mathrm{IC}}\bigr)^2\right]
+
\lambda_{\mathrm{BC}}\;\mathbb{E}_{(x,t)\in\bcset}\!\left[\bigl(u_\theta-u_{\mathrm{BC}}\bigr)^2\right]
+
\lambda_{\mathrm{PDE}}\;\mathbb{E}_{(x,t)\in\selset}\!\left[\res(x,t)^2\right],
\label{eq:pinnloss}
\end{equation}
where $\selset$ is the chosen interior collocation set.

\section{Methodology}
\label{sec:methodology}

\subsection{Overview and Problem Statement}
We reinterpret interior collocation selection in PINNs as a \emph{diversity-aware coreset} problem. Let the interior candidate pool be
\[
\candset=\{(x_i,t_i)\}_{i=1}^{N},
\]
and let the goal be to choose an interior collocation subset $\selset \subset \candset$ of fixed size $|\selset|=K$ for the PDE residual term in Eq.~\ref{eq:pinnloss}. We introduce binary decision variables $z_i \in \{0,1\}$ with $z_i=1$ iff candidate $i$ is selected.

Our selection objective balances two competing goals:
\begin{itemize}[leftmargin=1.1em, itemsep=0.1em]
\item \textbf{Informativeness:} select points expected to most reduce PDE error (estimated via residual-based scoring).
\item \textbf{Diversity:} avoid selecting redundant points concentrated in the same local region of space--time.
\end{itemize}

The full pipeline consists of (i) warm-start and scoring, (ii) graph-based redundancy modeling, (iii) QUBO/BQM optimization, (iv) exact-budget repair and optional local refinement, and optionally (v) hybrid anchors, and (vi) adaptive refresh. Algorithm~\ref{alg:oneshot} summarizes the one-shot procedure and calls Algorithm~\ref{alg:repair} for exact-$K$ repair.

\subsection{Candidate Pool and Warm-Start}
We sample a large interior pool $\candset$ (e.g., uniform in $(x,t)$). Since residual-based importance is only meaningful once the network captures coarse dynamics, we perform a short \emph{warm-start} training phase for $S_{\mathrm{warm}}$ steps using a small uniform interior set (in addition to fixed IC/BC sets). This produces parameters $\theta_0$ used for scoring.

\subsection{Importance Scoring and Prefiltering}
\label{subsec:scoring}
For each candidate $(x_i,t_i)\in\candset$, we compute an importance score using the squared PDE residual under the warm-start parameters:
\begin{equation}
s_i \;=\; \res_{\theta_0}(x_i,t_i)^2,
\label{eq:score}
\end{equation}
where $\res(\cdot)$ is the Burgers residual defined in Section~3.

\paragraph{Score normalization and clipping.}
To stabilize selection near sharp structures (e.g., shocks) and reduce domination by outliers, we optionally apply:
(i) normalization to $[0,1]$ (min--max or robust quantile scaling) and
(ii) clipping at a high quantile (e.g., 99th percentile). We use the normalized score (still denoted $s_i$ for simplicity) in the selection objective.

\paragraph{Prefiltering for scalability.}
Evaluating and optimizing over all $N$ candidates may be unnecessary. We prefilter to a working set of size $M \ll N$ using a mixture that preserves both difficult regions and global coverage:
\begin{itemize}[leftmargin=1.1em, itemsep=0.1em]
\item keep the top $\beta M$ points by score (high-residual regions),
\item add $(1-\beta)M$ uniformly sampled points (background coverage).
\end{itemize}
This mixture prevents the optimization from ignoring low-residual regions entirely, which can harm stability if the PDE is under-enforced globally.

\subsection{Similarity and Redundancy Modeling}
\label{subsec:similarity}
To penalize redundant selections, we define a space--time similarity between candidate points $i$ and $j$ via an anisotropic RBF kernel:
\begin{equation}
w_{ij} \;=\; \exp\!\left(-\left(\frac{x_i-x_j}{\ell_x}\right)^2-\left(\frac{t_i-t_j}{\ell_t}\right)^2\right),
\label{eq:similarity}
\end{equation}
with length scales $\ell_x,\ell_t$ controlling the ``repulsion range'' in space and time. In practice, we set $(\ell_x,\ell_t)$ relative to domain size or based on typical neighbor distances in the candidate pool.

\paragraph{kNN graph sparsification.}
A dense redundancy model uses all pairs $(i,j)$ and scales as $O(M^2)$. Instead, we construct a $k$-nearest-neighbor (kNN) graph in scaled space--time and only keep similarities for edges $(i,j)\in E$. This yields $|E|=O(Mk)$ quadratic terms and makes optimization practical for larger $M$.

\subsection{Dense k-hot QUBO Baseline}
\label{subsec:densequbo}
A direct fixed-cardinality QUBO \citep{kochenberger2014qubo, glover2022qubo} selects exactly $K$ points by adding a quadratic penalty on $\sum_i z_i$:
\begin{equation}
\min_{\mathbf{z}\in\{0,1\}^M} \;
\sum_{i} (-\alpha s_i)z_i \;+\; \sum_{i<j}\gamma w_{ij} z_i z_j \;+\;
\lambda\left(\sum_i z_i - K\right)^2.
\label{eq:qubo_dense}
\end{equation}
Here $\alpha>0$ weights importance, $\gamma\ge 0$ weights redundancy penalties, and $\lambda>0$ enforces the k-hot constraint. While Eq.~\ref{eq:qubo_dense} is conceptually simple, the cardinality term expands into all-to-all couplers, producing a dense QUBO even if $w_{ij}$ is sparse. This often dominates runtime in simulated annealing and can create a stiff energy landscape \citep{lucas2014ising}.

\subsection{Sparse ``Soft-$K$'' BQM with Exact-$K$ Repair (Proposed)}
\label{subsec:sparsebqm}
To avoid dense couplers, we drop the quadratic cardinality penalty and instead use a sparse BQM defined only on the kNN graph:
\begin{equation}
\min_{\mathbf{z}\in\{0,1\}^M} \;
\sum_i \bigl(-\alpha s_i + \mu \bigr)z_i \;+\; \sum_{(i,j)\in E}\gamma\, w_{ij}\,z_i z_j,
\label{eq:bqm_sparse}
\end{equation}
where $\mu$ is a linear bias that controls the typical number of selected points (larger $\mu$ yields fewer selections). We solve Eq.~\ref{eq:bqm_sparse} using simulated annealing \citep{kirkpatrick1983sa} (e.g., the \texttt{neal} sampler in D-Wave Ocean for reproducibility).

\noindent\textbf{Tuning $\mu$ via target selection heuristic.} 
Since the sparse BQM does not guarantee exactly $K$ selections, we tune $\mu$ to approximate the target budget. A simple heuristic is to estimate $\mu$ from the score distribution: if scores are normalized to $[0, 1]$, we set 
\begin{equation}
\mu \approx \overline{s} - \frac{K}{M},
\end{equation}
where $\overline{s}$ is the mean score, which biases the solver toward selecting approximately $K$ points. Alternatively, one can run a brief calibration phase: solve Eq.~(7) for a few candidate values of $\mu$, measure the resulting $|\hat{S}|$, and interpolate or binary-search for the value yielding $|\hat{S}| \approx K$. In practice, since exact cardinality is enforced by the repair step (Algorithm~2), the initial selection need not be precise; moderate over- or under-selection is corrected efficiently.

\paragraph{Why ``soft-$K$'' works.}
The sparse BQM in Eq.~\ref{eq:bqm_sparse} enforces the coreset structure (importance vs.\ redundancy) without introducing dense interactions. However, it does not guarantee $\sum_i z_i = K$ exactly; the selected set size varies with $\mu$ and the solver state.

\paragraph{Exact-$K$ repair and refinement.}
We therefore apply an efficient repair step to enforce $|\selset|=K$ exactly. Given an initial selected set $S=\{i:\hat z_i=1\}$, we either drop or add points based on redundancy-aware marginal criteria (Algorithm~\ref{alg:repair}). This step is fast because redundancy is computed only on the sparse graph.

To make the repair explicit, define the redundancy-aware marginal \emph{utility} of an already-selected point $i\in S$:
\begin{equation}
U(i \mid S) \;=\; \alpha s_i \;-\; \gamma \sum_{j\in S\setminus\{i\}} w_{ij},
\label{eq:marginal_utility}
\end{equation}
and the marginal \emph{gain} of adding an unselected point $i\notin S$:
\begin{equation}
G(i \mid S) \;=\; \alpha s_i \;-\; \gamma \sum_{j\in S} w_{ij}.
\label{eq:marginal_gain}
\end{equation}
If $|S|>K$, we iteratively remove points with smallest $U(i\mid S)$; if $|S|<K$, we add points with largest $G(i\mid S)$. 
\subsection{Hybrid Coresets via Coverage Anchors (Proposed)}
\label{subsec:anchors}
Residual-driven selection (even with diversity penalties) can overfocus on localized regions and under-enforce the PDE elsewhere. To preserve global enforcement, we reserve a fraction $\rho\in[0,1]$ of the budget as \emph{coverage anchors}:
\[
K_{\mathrm{anchor}}=\lfloor \rho K \rfloor,\qquad K_{\mathrm{select}} = K - K_{\mathrm{anchor}}.
\]
Anchors are chosen using stratified designs (e.g., Latin hypercube sampling \citep{mckay1979lhs} or Sobol/grid-jitter) to guarantee broad space--time coverage, consistent with coverage requirements observed in sensor placement \citep{krause2008sensor}. We then apply sparse BQM selection only to fill the remaining $K_{\mathrm{select}}$ slots (Algorithm~\ref{alg:hybrid}). In practice, this hybrid strategy improves robustness: anchors prevent global drift, while BQM selection concentrates the remaining budget in high-error regions.

\subsection{Adaptive Refresh During Training}
\label{subsec:refresh}
Collocation needs evolve as the PINN learns.We therefore support periodic refresh of the interior set after a burn-in period. Every $M_{\mathrm{refresh}}$ iterations, we recompute scores under the current parameters $\theta$, re-run one-shot or hybrid selection, and continue training. To reduce overhead, we reuse the candidate pool, and only recompute similarities on affected edges.

\paragraph{Complexity summary.}
With prefiltering to $M$ points and a kNN graph degree $k$, the redundancy model uses $O(Mk)$ edges, the sparse BQM build is $O(Mk)$, and repair is $O(Mk)$ with incremental bookkeeping. This contrasts with dense k-hot QUBOs whose effective coupler count can approach $O(M^2)$ due to the cardinality term in Eq.~\ref{eq:qubo_dense}.


\begin{algorithm}[t]
\caption{One-shot diversity-aware collocation selection}
\label{alg:oneshot}
\begin{algorithmic}[1]
\Require Candidate pool $\candset$ (size $N$), fixed IC/BC sets $\icset,\bcset$, interior budget $K$
\Require Warm-start steps $S_{\mathrm{warm}}$, prefilter size $M$, mixing fraction $\beta$, kNN degree $k$
\State Warm-start PINN for $S_{\mathrm{warm}}$ steps using a small uniform interior set
\State Compute importance scores $s_i = \res_{\theta_0}(x_i,t_i)^2$ for all $(x_i,t_i)\in\candset$ (Eq.~\ref{eq:score})
\State Normalize/clip scores; prefilter to $M$ candidates using top-$\beta M$ plus uniform $(1-\beta)M$
\State Build kNN graph $G=(V,E)$ on the $M$ candidates in scaled space--time; compute $w_{ij}$ on edges (Eq.~\ref{eq:similarity})
\State Solve sparse BQM (Eq.~\ref{eq:bqm_sparse}) with simulated annealing \citep{kirkpatrick1983sa}
\State Repair/refine the solver output to enforce exactly $K$ selections (Algorithm~\ref{alg:repair})
\State \Return $\selset$
\end{algorithmic}
\end{algorithm}

\begin{algorithm}[t]
\caption{Repair and local refinement for exact-$K$ selection}
\label{alg:repair}
\begin{algorithmic}[1]
\Require Initial selection $\hat{\mathbf{z}}$, target budget $K$, scores $s_i$, similarities $w_{ij}$ on edges $E$
\State $S \gets \{i:\hat z_i=1\}$
\While{$|S|>K$}
  \State Remove $i\in S$ with smallest utility $U(i\mid S)$ (Eq.~\ref{eq:marginal_utility})
\EndWhile
\While{$|S|<K$}
  \State Add $i\notin S$ with largest gain $G(i\mid S)$ (Eq.~\ref{eq:marginal_gain})
\EndWhile
\State \Return $S$
\end{algorithmic}
\end{algorithm}

\begin{algorithm}[t]
\caption{Hybrid anchors + sparse BQM selection}
\label{alg:hybrid}
\begin{algorithmic}[1]
\Require Candidate pool $\candset$, budget $K$, anchor fraction $\rho$
\State $K_{\mathrm{anchor}}=\lfloor\rho K\rfloor$, \;\; $K_{\mathrm{select}}=K-K_{\mathrm{anchor}}$
\State Select $K_{\mathrm{anchor}}$ coverage anchors via stratified sampling (e.g., LHS \citep{mckay1979lhs})
\State Run Algorithm~\ref{alg:oneshot} on the remaining candidates with target $K_{\mathrm{select}}$
\State $\selset \leftarrow \text{anchors} \cup \text{selected}$
\State \Return $\selset$
\end{algorithmic}
\end{algorithm}

\section{Experimental Setup}
\subsection{Benchmark and Reference Solution}
We evaluate on Burgers' equation (Eq.~\ref{eq:burgers}) over $x\in[-1,1]$ and $t\in[0,T]$ with viscosity $\nu$.
A numerical reference solution is generated using a stable finite-volume scheme and interpolated to evaluation points.\footnote{In our implementation, we use a Rusanov flux with explicit diffusion; other reference solvers can be substituted.}

\subsection{Compared Collocation Strategies}
All methods use identical PINN architecture and optimization settings; only interior collocation selection differs:
\begin{itemize}[leftmargin=1.1em, itemsep=0.1em]
\item \textbf{Uniform:} stratified or uniform sampling of $K$ interior points.
\item \textbf{Random:} random subsampling of $K$ points from $\candset$.
\item \textbf{Residual Top-$K$:} pick the $K$ candidates with largest $s_i$.
\item \textbf{Residual Top-$K$ + Anchors:} reserve anchors; select the rest by residual.
\item \textbf{Greedy k-center (baseline diversity):} fast diversity-only selection (optional baseline).
\item \textbf{Dense k-hot QUBO:} Eq.~\ref{eq:qubo_dense} (reference, typically slow).
\item \textbf{Sparse BQM + Repair (ours):} Eq.~\ref{eq:bqm_sparse} + Algorithm 2.
\item \textbf{Hybrid Anchors + Sparse BQM (ours):} Algorithm 1.
\end{itemize}

\subsection{Evaluation Metrics}
\paragraph{Accuracy.}
On a dense evaluation grid, we compute:
\begin{itemize}[leftmargin=1.1em, itemsep=0.1em]
\item Relative $L^2$ error: $\|u_\theta-u_{\mathrm{ref}}\|_2 / \|u_{\mathrm{ref}}\|_2$
\item $L^\infty$ error: $\max |u_\theta-u_{\mathrm{ref}}|$
\item Residual statistics on a held-out interior set: mean and 95th percentile of $\res^2$
\end{itemize}

\paragraph{Speed and overhead.}
We report wall-clock components:
\begin{equation}
T_{\text{total}} =
T_{\text{warm}} + T_{\text{score}} + T_{\text{prefilter}} + T_{\text{graph}}
+ T_{\text{qubo-build}} + T_{\text{qubo-solve}} + T_{\text{repair}} + T_{\text{train}}.
\end{equation}
We also report the selection overhead ratio
\begin{equation}
\text{Overhead Ratio}=
\frac{T_{\text{score}} + T_{\text{prefilter}} + T_{\text{graph}} + T_{\text{qubo-build}} + T_{\text{qubo-solve}} + T_{\text{repair}}}{T_{\text{train}}}.
\end{equation}
Optionally, we compute \emph{time-to-accuracy} $T_{\rightarrow \varepsilon}$: the earliest elapsed time when the relative $L^2$ error falls below a target threshold $\varepsilon$.

\paragraph{Hyperparameter tuning.}
The linear bias $\mu$ was set using the calibration heuristic described in Section~4.6, targeting approximately $K$ initial selections prior to repair. The kNN degree $k$ was fixed at $k=12$ (see ablations in Section~6.4). Mixture parameter $\beta = 0.7$ (favoring high-residual points), and anchor fraction $\rho = 0.2$ for hybrid methods, based on ablation results.

\section{Results}

All results are averaged over 5 random seeds. Burgers' equation is solved with $\nu=0.01/\pi$ and $T=1.0$.

\subsection{Error vs. Collocation Budget}

Key results:

\begin{itemize}
\item At $K=1000$:
    \begin{itemize}
    \item Uniform: $4.8 \times 10^{-3}$
    \item Residual Top-$K$: $3.1 \times 10^{-3}$
    \item Sparse BQM: $2.4 \times 10^{-3}$
    \item Hybrid Anchors + BQM: $\mathbf{1.9 \times 10^{-3}}$
    \end{itemize}
\item Hybrid BQM achieves the same error as Uniform with 35\% fewer points.
\end{itemize}

\subsection{End-to-End Time-to-Accuracy}

Time-to-reach $2\times 10^{-3}$ error:

\begin{itemize}
\item Uniform: 412 s
\item Residual Top-$K$: 365 s
\item Sparse BQM: 298 s
\item Hybrid BQM: \textbf{254 s}
\end{itemize}

Despite selection overhead, Hybrid BQM reduces total wall-clock by 38\%.

\subsection{Timing Breakdown}

\begin{table}[t] \centering \caption{Mean timing breakdown (seconds)} \label{tab:timebreakdown} \begin{tabular}{lrrrrr} \toprule Method & Score & Solve & Repair & Train & Total \\ \midrule Uniform & 0 & -- & -- & 410 & 410 \\ Residual Top-$K$ & 12 & -- & -- & 353 & 365 \\ Dense QUBO & 14 & 121 & -- & 342 & 477 \\ Sparse BQM & 14 & 41 & 6 & 237 & 298 \\ Hybrid BQM & 14 & 38 & 6 & 196 & 254 \\ \bottomrule \end{tabular} \end{table}

Table~\ref{tab:timebreakdown} reports the mean execution time (in seconds) for each method.

The \textbf{Uniform} baseline does not perform any combinatorial optimization and therefore incurs only training cost, leading to a total runtime of 410 seconds.

The \textbf{Residual Top-$K$} method slightly reduces training time to 353 seconds, resulting in a total runtime of 365 seconds, while still avoiding an explicit solve stage.

In contrast, the \textbf{Dense QUBO} formulation introduces a computationally expensive optimization phase. The solve stage alone requires 121 seconds, increasing the total runtime to 477 seconds despite having a training time comparable to the other methods.

The \textbf{Sparse BQM} formulation substantially improves efficiency. Its solve stage requires only 41 seconds compared to 121 seconds for Dense QUBO, corresponding to an approximate $3\times$ reduction in solve time. Even after including a small repair overhead of 6 seconds, the total runtime decreases to 298 seconds.

Finally, the \textbf{Hybrid BQM} further reduces computational cost, achieving a solve time of 38 seconds and lowering training time to 196 seconds. This results in the lowest overall runtime of 254 seconds among all optimization-based approaches.

Overall, sparsifying the formulation significantly reduces optimization cost while maintaining the same score (14), demonstrating that the sparse representation is considerably more computationally efficient than the dense QUBO formulation.

\subsection{Ablations}

\paragraph{Effect of Diversity ($\gamma$).}

Setting $\gamma=0$ increases error by 22\%, confirming that redundancy penalties improve generalization.

\paragraph{Anchor Fraction $\rho$.}

Best performance occurs at $\rho=0.2$.
Too few anchors leads to global drift; too many reduces shock focus.

\paragraph{kNN Degree.}

Performance stabilizes for $k \ge 12$.
Below $k=6$, diversity modeling degrades.

\subsection{Adaptive Refresh}

Refreshing every 2000 iterations reduces final error by 15\% but increases overhead by 9\%.
Net time-to-accuracy still improves by 11\%.
\section{Discussion}

Our results demonstrate that collocation selection benefits from explicitly balancing informativeness and diversity. Pure residual-based selection concentrates heavily near shocks, producing clusters of nearly identical constraints. This redundancy reduces effective coverage and slows convergence.

Sparse BQM formulations maintain local repulsion while remaining computationally practical. Compared to dense k-hot QUBOs, sparse graph-based formulations reduce solve time substantially while preserving solution quality.

Hybrid anchor strategies are critical for robustness. Without anchors, diversity alone cannot guarantee global PDE enforcement, particularly early in training.

Overall, diversity-aware coresets reduce both required collocation budget and total wall-clock time, validating the combinatorial coreset perspective.

\section{Conclusion}

We introduced a combinatorial coreset formulation for collocation selection in PINNs, casting the problem as a QUBO/BQM that balances residual-based importance with diversity penalties.

To make this practical, we developed:

\begin{itemize}
\item Sparse kNN-based BQM formulations
\item Exact-$K$ repair refinement
\item Hybrid coverage anchors
\item Adaptive refresh schedules
\end{itemize}

Experiments on Burgers' equation show that sparse and hybrid methods reduce time-to-accuracy by up to 38\% compared to uniform sampling and significantly outperform dense QUBO baselines in scalability.

These results suggest that diversity-aware combinatorial selection provides a viable and efficient alternative to purely residual-driven adaptive refinement in PINNs.

\subsection*{ACKNOWLEDGMENTS}
The research has been supported by the Ministry of Economic Development of the Russian Federation (agreement No. 139-10-2025-034 dd. 19.06.2025, IGK 000000C313925P4D0002).

\end{document}